\documentclass[10pt,twocolumn,letterpaper]{article}

\usepackage{cvpr}      %

\usepackage{graphicx}
\usepackage{amsmath}
\usepackage{amssymb}
\usepackage{booktabs}
\usepackage[usenames,dvipsnames]{xcolor}
\usepackage{url}
\usepackage{enumitem}
\usepackage{xspace}
\usepackage{paralist}
\usepackage{multirow}

\usepackage[pagebackref,breaklinks,colorlinks]{hyperref}

\makeatletter

\newcommand{\Rmnum}[1]{\expandafter\@slowromancap\romannumeral #1@}
\makeatother
\newcommand{\pardev}[2]{\frac{\partial {#1}}{\partial {#2}}}

\usepackage{bm}

\newif\ifshowcomments
\showcommentstrue %

\ifshowcomments
    \newcommand{\todo}[1]{\noindent\textcolor{BrickRed}{\textbf{[TODO:~}#1\textbf{]}}}
    \newcommand{\otmar}[1]{\noindent\textcolor{ForestGreen}{\textbf{[Otmar:~}#1\textbf{]}}}
    \newcommand{\xu}[1]{\noindent\textcolor{Salmon}{\textbf{[Xu:~}#1\textbf{]}}}
    \newcommand{\jie}[1]{\noindent\textcolor{Dandelion}{\textbf{[Jie:~}#1\textbf{]}}}
    \newcommand{\AS}[1]{\noindent\textcolor{BrickRed}{\textbf{[Adrian:~}#1\textbf{]}}}
    \newcommand{\MK}[1]{\noindent\textcolor{Dandelion}{\textbf{[Manuel:~}#1\textbf{]}}}
    \newcommand{\OH}[1]{{\color{blue}[OH: #1]}}
    \newcommand{\pmnote}[1]{\PM{#1}}
    \newcommand{\oh}[1]{\OH{#1}}
    \newcommand{\cg}[1]{{\color{magenta}[CG: #1]}}
    \newcommand{\JZ}[1]{{\color{red}[JZ: #1]}}
\else
    \newcommand{\todo}[1]{\unskip}
    \newcommand{\otmar}[1]{\unskip}
    \newcommand{\xu}[1]{\unskip}
    \newcommand{\jie}[1]{\unskip}
    \newcommand{\AS}[1]{\unskip}
    \newcommand{\MK}[1]{\unskip}
    \newcommand{\OH}[1]{\unskip}
    \newcommand{\pmnote}[1]{\unskip}
    \newcommand{\oh}[1]{\unskip}
    \newcommand{\cg}[1]{\unskip}
    \newcommand{\JZ}[1]{\unskip}

\fi    
    
\newcommand{\methodname}{Vid2Avatar\xspace}
\newcommand{\suppmat}{Supp. Mat\xspace}

\newcommand{\figref}[1]{Fig.~\ref{#1}}
\newcommand{\tabref}[1]{Tab.~\ref{#1}}
\newcommand{\secref}[1]{Sec.~\ref{#1}}

\definecolor{babyblue}{rgb}{0.54, 0.81, 0.94}

\usepackage[capitalize]{cleveref}
\crefname{section}{Sec.}{Secs.}
\Crefname{section}{Section}{Sections}
\Crefname{table}{Table}{Tables}
\crefname{table}{Tab.}{Tabs.}

\def\cvprPaperID{1072} 
\def\confName{CVPR}
\def\confYear{2023}

\begin{document}

\title{\methodname: 3D Avatar Reconstruction from Videos in the Wild \\ via Self-supervised Scene Decomposition}

\author{Chen Guo$^{1}$ \quad Tianjian Jiang$^{1}$ \quad Xu Chen$^{1,2}$ \quad Jie Song$^{1}$ \quad Otmar Hilliges$^{1}$ \\
 $^1$ETH Z{\"u}rich \quad 
 $^2$Max Planck Institute for Intelligent Systems, T{\"u}bingen \\
}

\twocolumn[{%
\renewcommand\twocolumn[1][]{#1}%
\maketitle

\vspace{-3em}
\begin{center}
    \captionsetup{type=figure}
   \includegraphics[width=\linewidth,trim=10 5 0 0,clip]{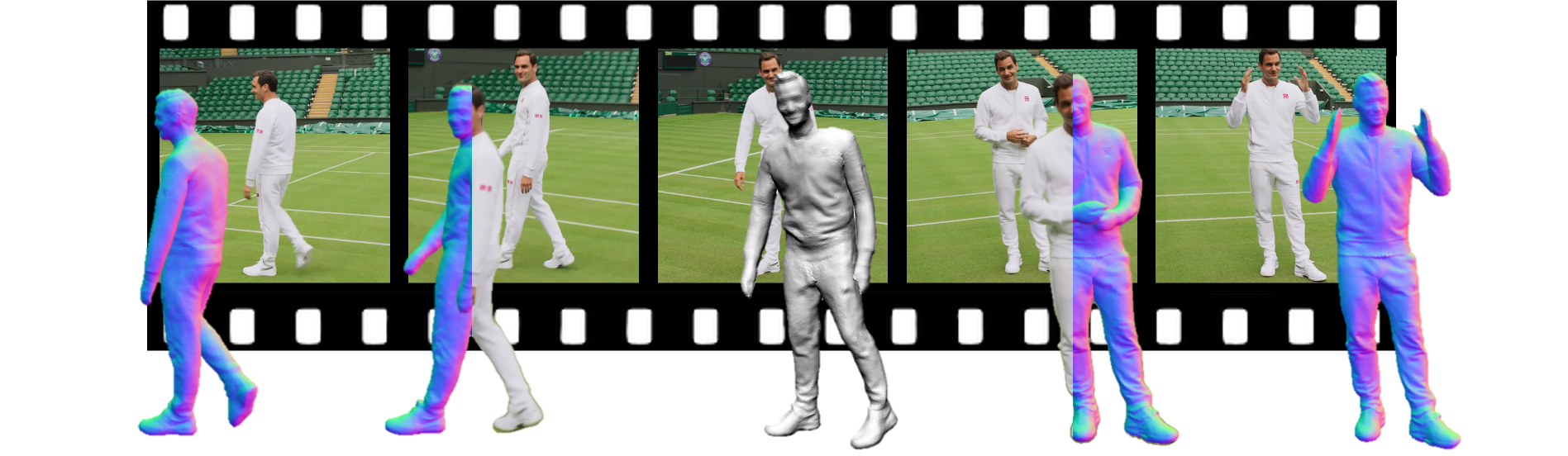}
    \captionof{figure}{We propose \methodname, a method to reconstruct detailed 3D avatars from monocular videos in the wild via self-supervised scene decomposition. Our method does not require any groundtruth supervision or priors extracted from large datasets of clothed human scans, nor do we rely on any external segmentation modules.
    }\label{fig:teaser}

\end{center}%

}]


\newcommand{\figurePipeline}{

\begin{figure*}[t]
\includegraphics[width=\linewidth, trim=0 5 0 0,clip]{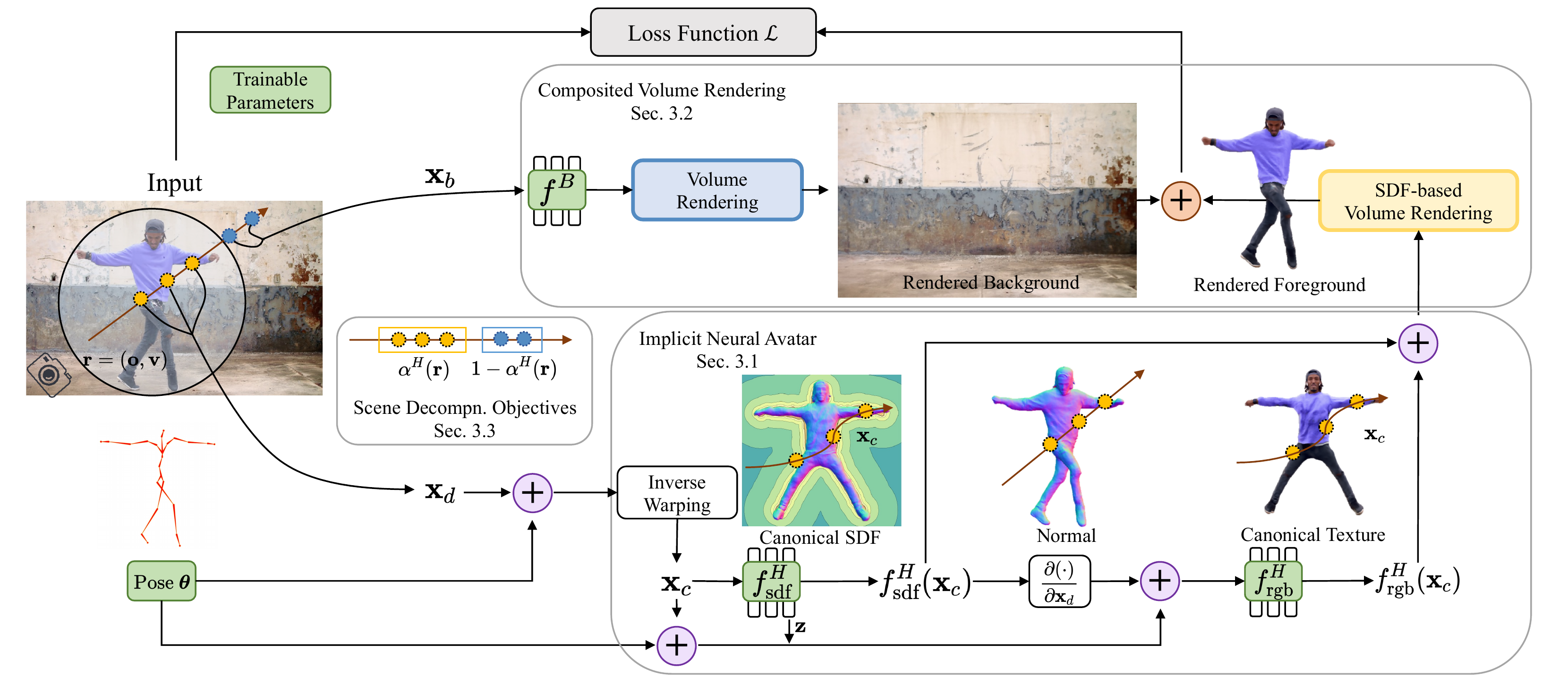}
\caption{\textbf{Method Overview.} Given a ray $\mathbf{r}$ with camera center $\mathbf{o}$ and ray direction $\mathbf{v}$, we sample points densely ($\mathbf{x}_d$) and coarsely ($\mathbf{x}_b$) along the ray for the spherical inner volume and outer volume respectively. Within the foreground sphere, we warp all sampled points into canonical space via inverse warping and evaluate the SDF of the canonical correspondences $\mathbf{x}_c$ via the canonical shape network $f_{\text{sdf}}^{H}$. We calculate the spatial gradient of the sampled points in deformed space and concatenate them with the canonical points $\mathbf{x}_c$, the pose parameters $\boldsymbol{\theta}$, and the extracted geometry feature vectors $\mathbf{z}$ to form the input to canonical texture network $f_\text{rgb}^{H}$ which predicts color values for $\mathbf{x}_c$. We apply surface-based volume rendering for the dynamic foreground and standard volume rendering for the background, and then composite the foreground and background components to attain the final pixel color. We minimize the loss $\mathcal{L}$ that compares the color predictions with the image observations along with novel scene decomposition objectives.}
\label{fig:pipeline}
\end{figure*}
}

\newcommand{\figurejointopt}{

\begin{figure}[t]
\includegraphics[width=\linewidth,trim=0 10 0 0,clip]{figures/jointopt.png}

\caption{\textbf{Importance of jointly pose optimization.} Jointly pose optimization corrects initial pose estimates and achieves better reconstruction quality.}

\label{fig:jointopt}
\end{figure}
} 

\newcommand{\figurewobg}{

\begin{figure}[t]
\includegraphics[width=\linewidth,trim=0 15 0 0,clip]{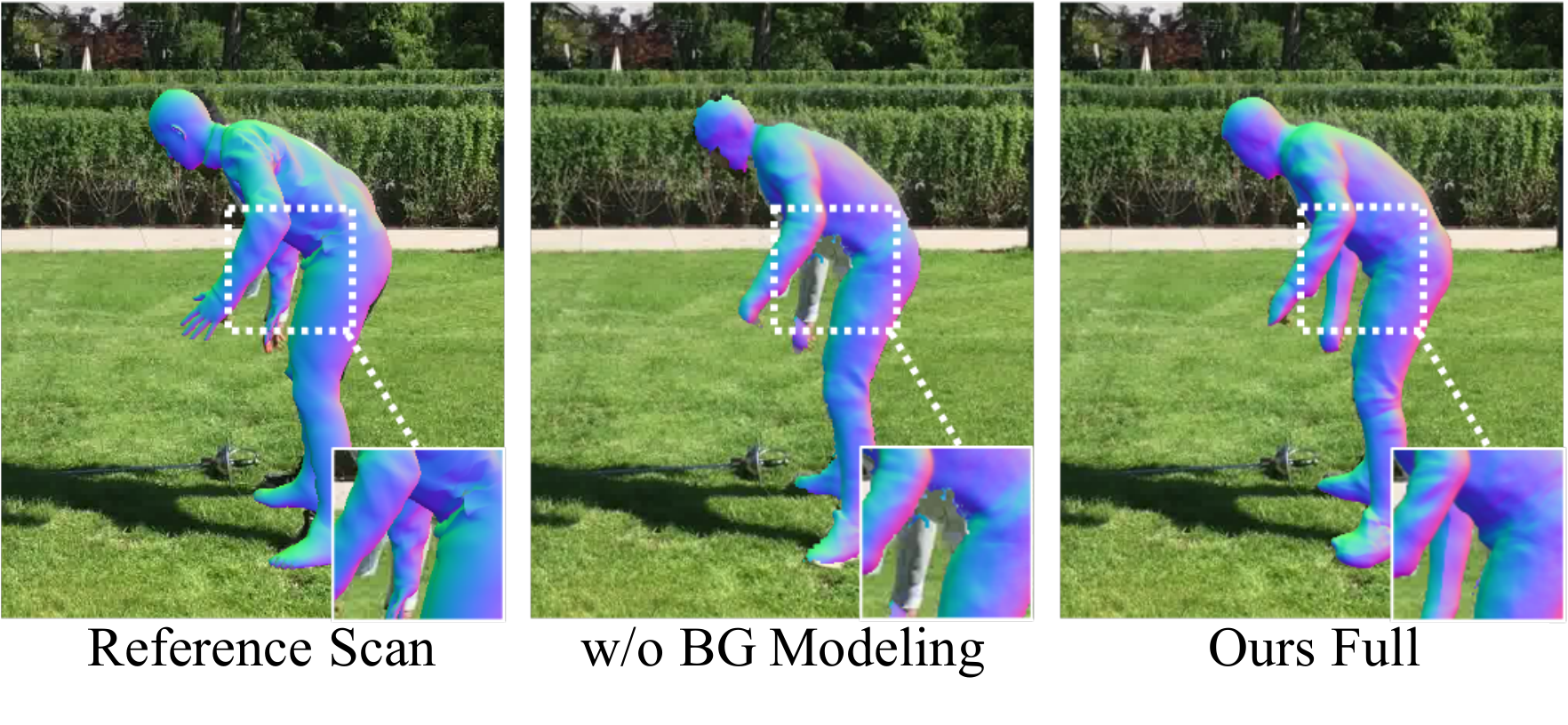}

\caption{\textbf{Importance of modeling background.} Without modeling the background, the decoupling between the human and the background is upper-bounded by the off-the-shelf segmentation tool and only yields worse reconstruction results.}

\label{fig:wobg}
\end{figure}
} 

\newcommand{\figuredeepcap}{

\begin{figure}[t]
\includegraphics[width=\linewidth,trim=0 10 0 0,clip]{figures/deepcap.pdf}

\caption{\textbf{Qualitative comparison with DeepCap.} Our method recovers better dynamic surface details (e.g., cloth wrinkles) and realistic facial features.}

\label{fig:deepcap}
\end{figure}
} 

\newcommand{\figuremonoperfcap}{

\begin{figure}[t]
\includegraphics[width=\linewidth,trim=0 10 0 0,clip]{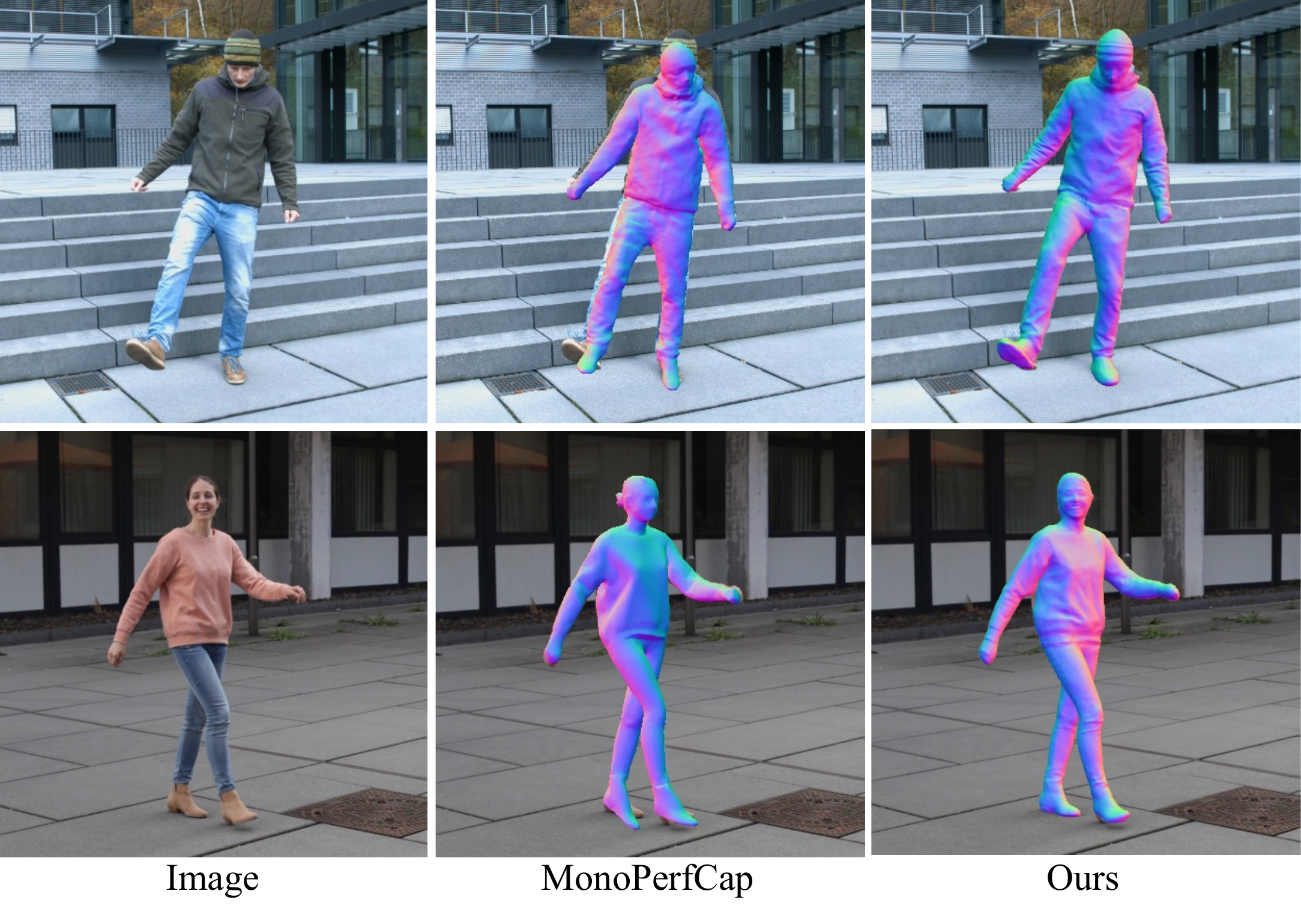}

\caption{\textbf{Qualitative comparison with MonoPerfCap.} Our method recovers better dynamic surface details (e.g., cloth wrinkles) and realistic facial features.}

\label{fig:monoperfcap}
\end{figure}
}

\newcommand{\figuredensityreg}{

\begin{figure}[t]

\includegraphics[width=\linewidth,trim=0 10 0 0,clip]{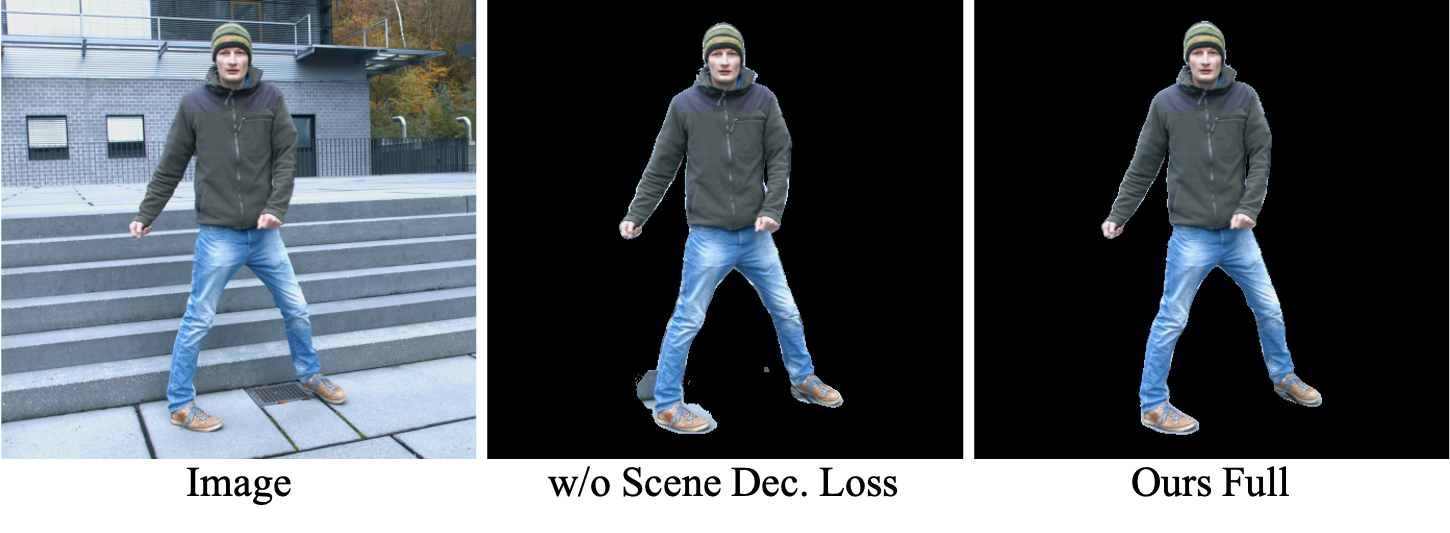}

\caption{\textbf{Importance of scene decomposition loss.} Without the scene decomposition loss, the segmentation includes undesirable background parts due to similar pixel values across discontinuities.}
\vspace{-1em}
\label{fig:densityreg}
\end{figure}
}

\newcommand{\figureneuman}{

\begin{figure}[t]
\includegraphics[width=\linewidth,trim=0 10 0 0,clip]{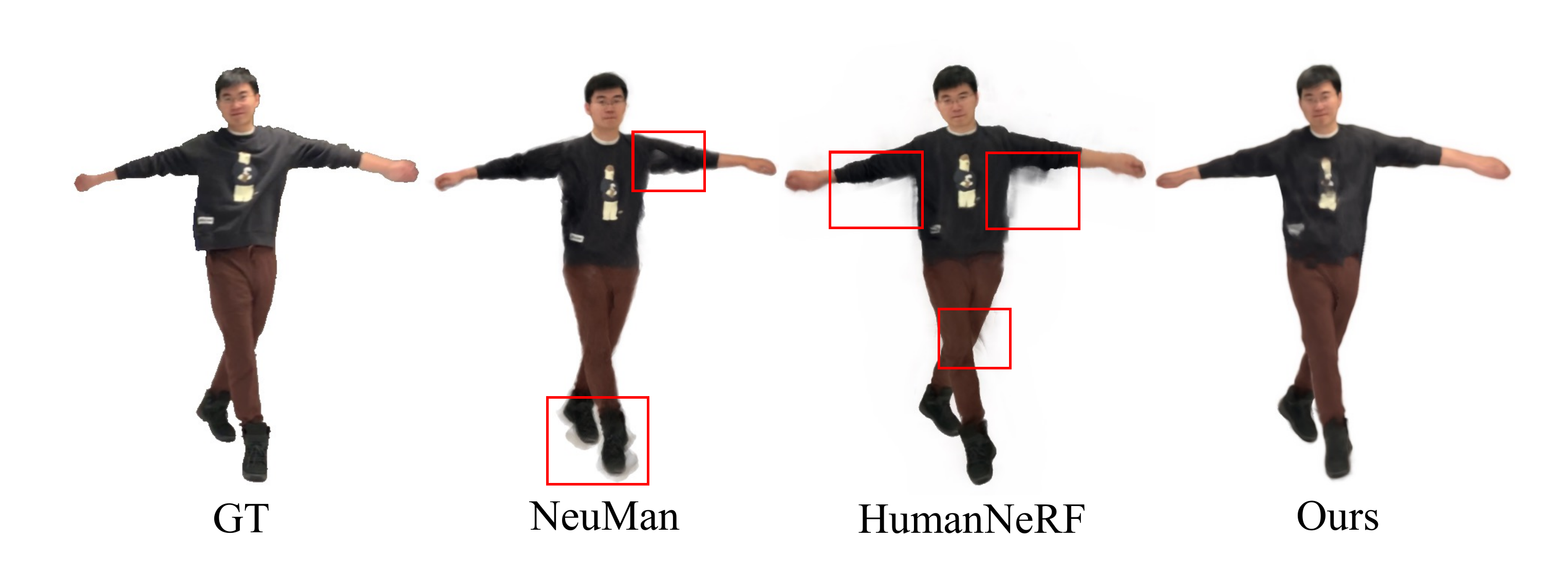}

\caption{\textbf{Qualitative view synthesis comparison.} Our method achieves comparable and even better novel view synthesis results compared to NeRF-based methods (see also \secref{sec:vs}).
}

\label{fig:neuman}
\end{figure}
} 

\newcommand{\figuremask}{

\begin{figure}[t]
\includegraphics[width=\linewidth,trim=0 15 0 0,clip]{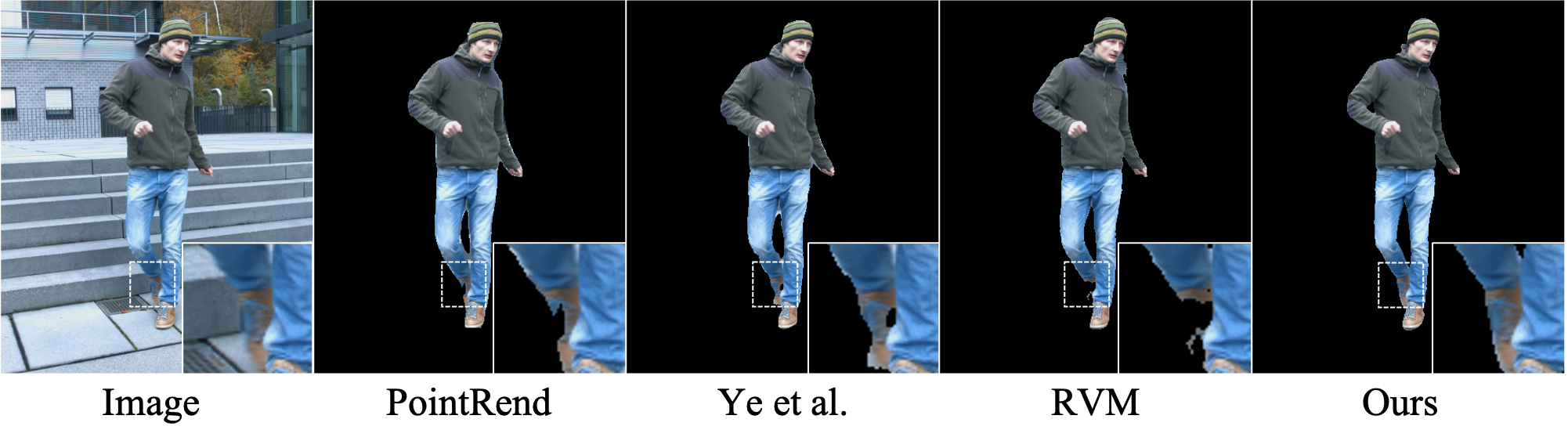}

\caption{\textbf{Qualitative mask comparison.}
Our method generates more detailed and robust segmentations compared to 2D segmentation methods by incorporating 3D knowledge.
}

\label{fig:mask}
\end{figure}
}

\newcommand{\figurerecon}{

\begin{figure*}[t]
\includegraphics[width=\linewidth,trim=0 10 0 0,clip]{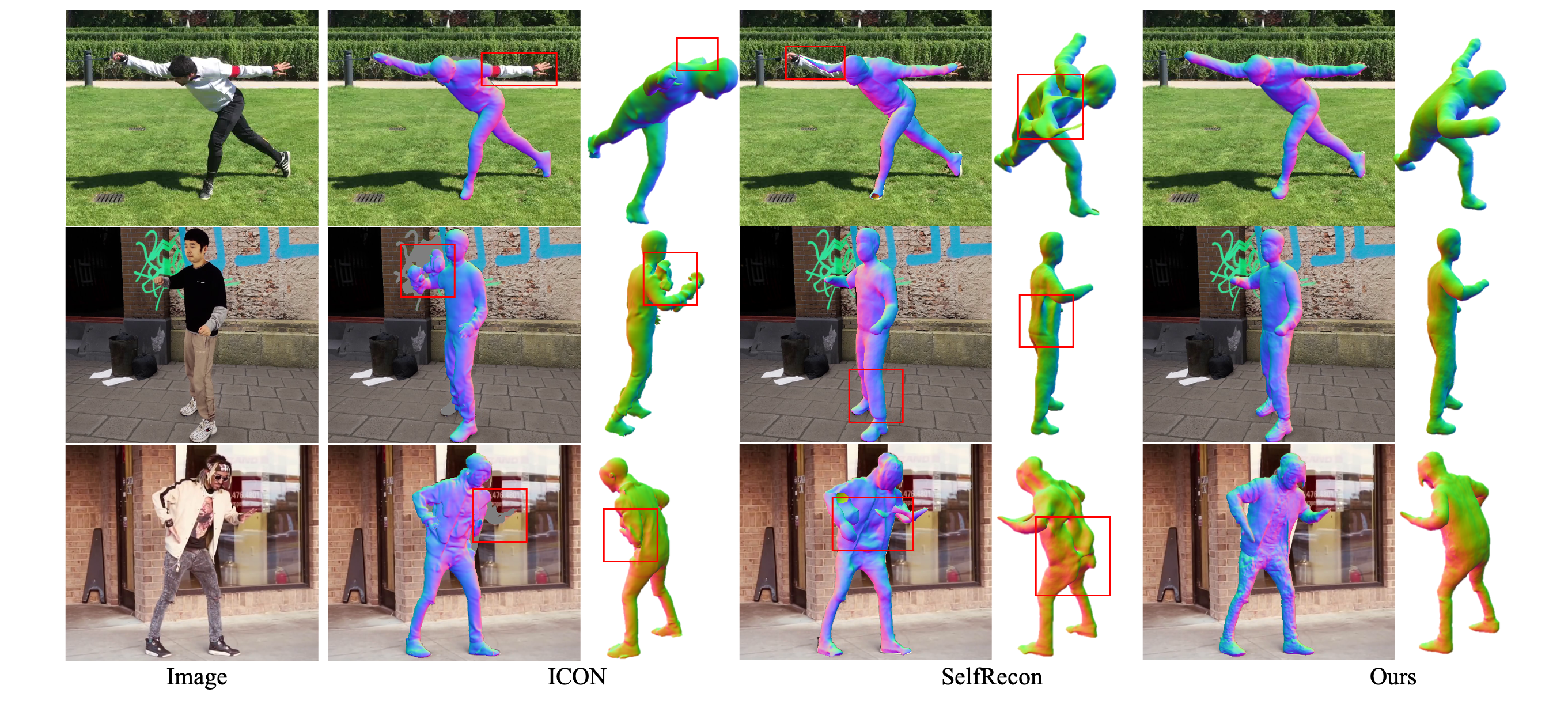}

\caption{\textbf{Qualitative reconstruction comparison.} Data source top to bottom: 3DPW, SynWild, Online. ICON and SelfRecon produce less detailed and even physically implausible reconstructions (incomplete human bodies). In contrast, our method generates complete human bodies and achieves a detailed (e.g., cloth wrinkles) and temporally consistent shape reconstruction.}

\label{fig:recon}
\end{figure*}
}

\newcommand{\figuredemo}{

\begin{figure*}[t]
\includegraphics[width=\linewidth,trim=0 2 0 0,clip]{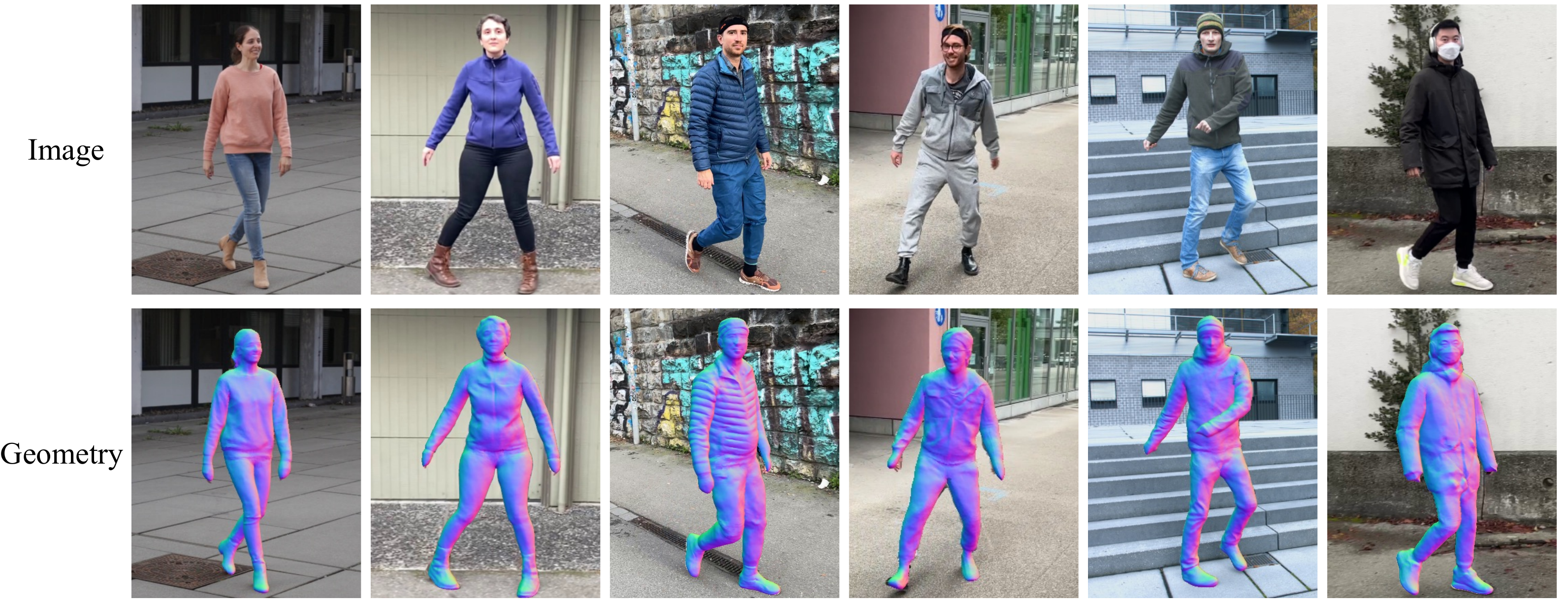}
\vspace{-1em}
\caption{\textbf{Qualitative results.} We show qualitative results of our method from monocular in-the-wild videos.}

\label{fig:demo}
\end{figure*}
}
  
\newcommand{\figurescenedecomp}{

\begin{figure*}[t]
\includegraphics[width=0.95\linewidth,trim=0 2 0 0,clip]{figures/scenedecomp.pdf}
\vspace{-1em}
\caption{\textbf{Qualitative results of background rendering.} We show qualitative results of our modeled backgrounds. }

\label{fig:scenedecomp}
\end{figure*}
}

\newcommand{\figureanimation}{

\begin{figure}[t]
\includegraphics[width=\linewidth,trim=0 12 0 0,clip]{figures/animation.pdf}
\vspace{-1em}
\caption{\textbf{Animation results.} Given driving signals from different sources, our reconstructed 3D avatars can be animated to novel new poses and can be rendered along with the original background.}

\label{fig:animation}
\end{figure}
}

\newcommand{\figuresuppneuman}{

\begin{figure}[t]
\includegraphics[width=\linewidth,trim=0 12 0 0,clip]{figures/supp_neuman.pdf}
\vspace{-1em}
\caption{\textbf{Additional qualitative view synthesis comparison.} Our method achieves comparable and even better novel view synthesis results compared to NeRF-based methods.}

\label{fig:suppneuman}
\end{figure}
}

\newcommand{\figureposefailure}{

\begin{figure}[t]
\includegraphics[width=\linewidth,trim=0 12 0 0,clip]{figures/posefailure.pdf}
\vspace{-1em}
\caption{\textbf{Failure case.} Unreasonable pose initialization leads to the incorrect reconstruction, especially when the RGB information is not enough for the pose correction.}

\label{fig:posefailure}
\end{figure}
}

\newcommand{\figuregtrecon}{

\begin{figure}[t]
\includegraphics[width=\linewidth,trim=0 2 0 0,clip]{figures/gtrecon.pdf}
\vspace{-1em}
\caption{\textbf{SynWild Dataset.} We show sample images and their corresponding ground-truth meshes from the SynWild dataset.}

\label{fig:gtrecon}
\end{figure}
}

\newcommand{\figureaddquali}{

\begin{figure}[t]
\includegraphics[width=\linewidth,trim=0 10 0 0,clip]{figures/additional_quali.pdf}
\vspace{-1em}
\caption{\textbf{Additional qualitative results.} We show qualitative results of our method from monocular in-the-wild videos.}

\label{fig:addquali}
\end{figure}
}


\begin{abstract}
\vspace{-1.2em}

\setlength{\topsep}{0pt}
\setlength{\parskip}{.5ex}
\renewcommand{\floatsep}{1ex}
\renewcommand{\textfloatsep}{1ex}
\renewcommand{\dblfloatsep}{1ex}
\renewcommand{\dbltextfloatsep}{1ex}

We present \methodname, a method to learn human avatars from monocular in-the-wild videos. 
Reconstructing humans that move naturally from monocular in-the-wild videos is difficult. Solving it requires accurately separating humans from arbitrary backgrounds.
Moreover, it requires reconstructing detailed 3D surface from short video sequences, making it even more challenging.
Despite these challenges, our method does not require any groundtruth supervision or priors extracted from large datasets of clothed human scans, nor do we rely on any external segmentation modules.
Instead, it solves the tasks of scene decomposition and surface reconstruction directly in 3D by modeling both the human and the background in the scene jointly, parameterized via two separate neural fields.
Specifically, we define a temporally consistent human representation in canonical space and formulate a global optimization over the background model, the canonical human shape and texture, and per-frame human pose parameters.
A coarse-to-fine sampling strategy for volume rendering and novel objectives are introduced for a clean separation of dynamic human and static background, yielding detailed and robust 3D human geometry reconstructions.
We evaluate our methods on publicly available datasets and show improvements over prior art. Project page: \href{https://moygcc.github.io/vid2avatar/}{https://moygcc.github.io/vid2avatar/}.

\end{abstract}


\section{Introduction}
Being able to reconstruct detailed avatars from readily available ``in-the-wild'' videos, for example recorded with a phone, would enable many applications in AR/VR, in human-computer interaction, robotics and in the movie and sports industry.
Traditionally, high-fidelity 3D reconstruction of dynamic humans has required calibrated multi-view systems \cite{4178157, journals/tvcg/LiuDX10, 1335229, 10.1145/1360612.1360697, tsiminaki:hal-00977755, leroy:hal-02975479, collet2015msft}, which are expensive and require highly-specialized expertise to operate.
In contrast, emerging applications such as the Metaverse require more light-weight and practical solutions in order to make the digitization of humans a widely available technology. 
Reconstructing humans that move naturally from monocular in-the-wild videos is clearly a difficult problem.
Solving it requires accurately separating humans from arbitrary backgrounds, without any prior knowledge about the scene or the subject.
Moreover it requires reconstructing detailed 3D surface from short video sequences, made even more challenging due to depth ambiguities, the complex dynamics of human motion and the high-frequency surface details.

Traditional template-based approaches \cite{Xu:2018:MHP:3191713.3181973, 10.1145/3311970, deepcap} cannot generalize to in-the-wild settings due to the requirement for a pre-scanned template and manual rigging. Methods that are based on explicit mesh representations are limited to a fixed topology and resolution \cite{alldieck2018video, guo2021human, Moon_2022_ECCV_ClothWild, casado2022pergamo}. 
Fully-supervised methods that directly regress 3D surfaces from images~\cite{NEURIPS2020_690f44c8, saito2019pifu, saito2020pifuhd, huang2020arch, He_2021_ICCV, zheng2021pamir, xiu2022icon} struggle with difficult out-of-distribution poses and shapes, and do not always predict temporally consistent reconstructions.
Fitting neural implicit surfaces to videos has recently been demonstrated \cite{su2021anerf, peng2021neural, jiang2022selfrecon, zheng2022imavatar, su2022danbo, weng_humannerf_2022_cvpr, jiang2022neuman}. However, these methods depend on pre-segmented inputs and are therefore not robust to uncontrolled visual complexity and are upper-bounded in their reconstruction quality by the segmentation method.

In this paper, we introduce \methodname, a method to learn human avatars from monocular in-the-wild videos without requiring any groundtruth supervision or priors extracted from large datasets of clothed human scans, nor do we rely on any external segmentation modules.
We solve the tasks of scene separation and surface reconstruction directly in 3D. 
To achieve this, we model both the foreground (i.e., human) and the background in the scene implicitly, parameterized via two separate neural fields.
A key challenge is to associate 3D points to either of these fields without reverting to 2D segmentation. 
To tackle this challenge, our method builds-upon the following core concepts:
i) We define a single temporally consistent representation of the human shape and texture in canonical space and leverage the inverse mapping of a parametric body model to learn from deformed observations.
ii) A global optimization formulation jointly optimizes the parameters of the background model, the canonical human shape and its appearance, and the pose estimates of the human subject over the entire sequence. 
iii) A coarse-to-fine sampling strategy for volume rendering that naturally leads to a separation of dynamic foreground and static background. 
iv) Novel objectives that further improve the scene decomposition and lead to sharp boundaries between the human and the background, even when both are in contact (e.g., around the feet), yielding better geometry and appearance reconstructions.  

More specifically, we leverage an inverse-depth parameterization in spherical coordinates~\cite{kaizhang2020} to coarsely separate the static background from the dynamic foreground. 
Within the foreground sphere, we leverage a surface-guided volume rendering approach to attain densities via the conversion method proposed in~\cite{yariv2021volume}. Importantly, we warp all sampled points into canonical space and update the human shape field dynamically. 
To attain sharp boundaries between the dynamic foreground and the scene, we introduce two optimization objectives that encourage a quasi-discrete binary distribution of ray opacities and penalize non-zero opacity for rays that do not intersect with the human. 
The final rendering of the scene is then attained by differentiable composited volume rendering.

We show that this optimization formulation leads to clean scene decomposition and high-quality 3D reconstructions of the human subject. 
In detailed ablations, we shed light on the key components of our method. 
Furthermore, we compare to existing methods in 2D segmentation, novel view synthesis, and reconstruction tasks, showing that our method performs best across several datasets and settings. To allow for quantitative comparison across methods, we contribute a novel semi-synthetic test set that contains accurate 3D geometry of human subjects.
Finally, we demonstrate the ability to reconstruct different humans in detail from online videos and hand-held mobile phone video clips.

In summary, our contributions are:
\begin{compactitem}
 \item a method to reconstruct detailed 3D avatars from in-the-wild monocular videos via self-supervised scene decomposition; and
 \item to achieve robust and detailed 3D reconstructions of the human even under challenging poses and environments without requiring external segmentation methods; and
 \item a novel semi-synthetic testing dataset that for the first time allows comparing monocular human reconstruction methods on realistic scenes. The dataset contains rich annotations of the 3D surface.
\end{compactitem}

\noindent Code and data will be made available for research purposes.


\section{Related Work}

\vspace{-0.1cm}

\paragraph{Reconstructing Human from Monocular Video}
Traditional works for monocular human performance capture require personalized rigged templates as prior and track the pre-defined human model based on 2D observations \cite{Xu:2018:MHP:3191713.3181973, 10.1145/3311970, deepcap}. These works require pre-scanning of the performer and post-processing for rigging, preventing such methods from being deployed to real-life applications. Some methods attempt to save the need for pre-scanning and manual rigging \cite{alldieck2018video, guo2021human, Moon_2022_ECCV_ClothWild, casado2022pergamo}. However, the explicit mesh representation is limited to a fixed resolution and cannot represent details like the face. Regression-based methods that directly regress 3D surfaces from images have demonstrated compelling results \cite{NEURIPS2020_690f44c8, saito2019pifu, saito2020pifuhd, huang2020arch, He_2021_ICCV, zheng2021pamir, xiu2022icon, alldieck2022phorhum, li2022neurips}. However, they require high-quality 3D data for supervision and cannot maintain the space-time coherence of the reconstruction over the whole sequence. Recent works fit implicit neural fields to videos via neural rendering to obtain articulated human models \cite{su2021anerf, peng2021neural, jiang2022selfrecon, zheng2022imavatar, su2022danbo, weng_humannerf_2022_cvpr, jiang2022neuman}. HumanNeRF \cite{weng_humannerf_2022_cvpr} extends articulated NeRF to improve novel view synthesis. NeuMan \cite{jiang2022neuman} further adds a scene NeRF model. Both methods model the human geometry with a density field, only yielding a noisy, and often low-fidelity human reconstruction. SelfRecon \cite{jiang2022selfrecon} deploys neural surface rendering \cite{yariv2020multiview} to achieve consistent reconstruction over the sequence. However, all aforementioned methods rely on pre-segmented inputs and are therefore not robust to uncontrolled visual complexity and are upper-bounded in their reconstruction quality by the external segmentation method. In contrast, our method solves the tasks of scene decomposition and surface reconstruction jointly in 3D without using segmentation modules.
\vspace{-0.3cm}
\paragraph{Reconstructing Human from Multi-view/Depth} 
The high fidelity 3D reconstruction of dynamic humans has required calibrated dense multi-view systems \cite{4178157, journals/tvcg/LiuDX10, 1335229, 10.1145/1360612.1360697, tsiminaki:hal-00977755, leroy:hal-02975479, collet2015msft} which are expensive and laborious to operate and require highly-specialized expertise. Recent works \cite{peng2021animatable, 2021narf, xu2021hnerf, zhang2021stnerf, HVTR:3DV2022, ARAH:2022:ECCV, li2022tava, xu2022sanerf} attempt to reconstruct humans from more sparse settings by deploying neural rendering. Depth-based approaches \cite{newcombe2011kinectfusion, newcombe2015dynamicfusion, Bozic_2021_CVPR} reconstruct the human shape by fusing depth measurements across time. Follow-up work \cite{BodyFusion, DoubleFusion, li2021posefusion, burov2021dsfn, Dong_2022_CVPR} builds upon this concept by incorporating an articulated motion prior and a parametric body shape prior. While the aforementioned methods achieve compelling results, they still require a specialized capturing setup and are hence not applicable to in-the-wild settings. In contrast, our method recovers the dynamic human shape in the wild from a monocular RGB video as the sole input.

\paragraph{Moving Object Segmentation}
Traditional research in moving object segmentation has been extensively conducted at the image level (i.e. 2D). One line of research relies on motion clues to segment objects with different optical flow patterns \cite{10.1007/978-3-319-46484-8_26, 6751331, yang2021selfsupervised, 8953201, ye2022sprites}, while another line of work, termed video matting \cite{BMSengupta20, MODNet, rvm} is trained on videos with human-annotated masks to directly regress the alpha-channel values during inference. Those approaches are not without limitations, as they focus on image-level segmentation and incorporate no 3D knowledge. Thus, they cannot handle complicated backgrounds without enough color contrast between the human and the background. Recent works learn to decompose dynamic objects and the static background simultaneously in 3D by optimizing multiple NeRFs \cite{yuan2021star, tschernezki21neuraldiff, https://doi.org/10.48550/arxiv.2205.15838, https://doi.org/10.48550/arxiv.2207.11232}. Such methods perform well for non-complicated dynamic objects but are not directly applicable to articulated humans with intricate motions.


\section{Method}
\figurePipeline
We introduce \methodname, a method for detailed geometry and appearance reconstruction of implicit neural avatars from monocular videos in the wild. Our method is schematically illustrated in \figref{fig:pipeline}. Reconstructing humans from in-the-wild videos is clearly challenging. Solving it requires accurately segmenting humans from arbitrary backgrounds without any prior knowledge about the appearance of the scene or the subject and requires reconstructing detailed 3D surface and appearance from short video sequences. 
In contrast to prior works that utilize off-the-shelf 2D segmentation tools or manually labeled masks, we solve the tasks of scene decomposition and surface reconstruction directly in 3D. 
To achieve this, we model both the human and background in the scene implicitly, parameterized via two separate neural fields which are learned jointly from images to composite the whole scene. To alleviate the ambiguity of in-contact body and scene parts and to better delineate the surfaces, we contribute novel objectives that leverage the dynamically updated human shape in canonical space to regularize the ray opacity.

We parameterize the 3D geometry and texture of clothed humans as a pose-conditioned implicit signed-distance field (SDF) and texture field in canonical space (\secref{sec:avatar_model}). 
We then model the background using a separate neural radiance field (NeRF). The human shape and appearance fields alongside the background field are learned from images jointly via differentiable composited neural volume rendering (\secref{sec:volumerendering}). 
Finally, we leverage the dynamically updated canonical human shape to regularize the ray opacities (\secref{sec:scenedecompobj}). The training is formulated as global optimization to jointly optimize the dynamic foreground and static background fields, and the per-frame pose parameters (\secref{sec:globalopt}).

\subsection{Implicit Neural Avatar Representation}
\label{sec:avatar_model}

\paragraph{Canonical Shape Representation.}
We model the human shape in canonical space to form a single, temporally consistent representation and use a neural network $f_{\text{sdf}}^{H}$ to predict the signed distance value for any 3D point $\mathbf{x}_c$ in this space. To model pose-dependent local non-rigid deformations such as dynamically changing wrinkles on clothes, we concatenate the human pose $\boldsymbol{\theta}$ as an additional input and model $f_{\text{sdf}}^{H}$ as:
\begin{equation}
    f_{\text{sdf}}^{H}: \mathbb{R}^{3 + n_{\theta}} \rightarrow \mathbb{R}^{1 + 256} .
\end{equation}
The pose parameters $\boldsymbol{\theta}$ are defined analogously to SMPL~\cite{loper2015smpl}, with dimensionality $n_{\theta}$. Furthermore, $f_{\text{sdf}}^{H}$ outputs global geometry features $\textbf{z}$ of dimension 256. With slight abuse of notation, we also use $f_{\text{sdf}}^{H}$ to refer to the SDF value only. The canonical shape $\mathcal{S}$ is given by the zero-level set of $f_{\text{sdf}}^{H}$:
\begin{equation}
     \mathcal{S} = \{ \mathbf{\ x}_c \ |\ f_{\text{sdf}}^{H}(\mathbf{x}_c,\boldsymbol{\theta}) = 0 \ \} .
\end{equation}
\paragraph{Skeletal Deformation.} Given the bone transformation matrix $\mathbf{B}_i$ for joint $i \in \{1,...,n_b\}$ which are derived from the body pose $\boldsymbol{\theta}$, a canonical point $\mathbf{x}_c$ is mapped to the deformed point $\mathbf{x}_d$ via linear-blend skinning:

\begin{equation}
\label{eq:lbs}
    \mathbf{x}_d = \sum_{i = 1}^{n_b} w_{c}^i \mathbf{B}_i \, \mathbf{x}_c .
\end{equation}
The canonical correspondence $\mathbf{x}_{c}$ for points $\mathbf{x}_{d}$ in deformed space is defined by the inverse of Eq.~\ref{eq:lbs}:

\begin{equation}
\label{eq:lbs_inv}
    \mathbf{x}_c = (\sum_{i = 1}^{n_b} w_{d}^i \mathbf{B}_i)^{-1}\ \mathbf{x}_d 
\end{equation}
Here, $n_b$ denotes the number of bones in the transformation, and $\mathbf{w}_{(\cdot)} = \{w_{(\cdot)}^1,...,w_{(\cdot)}^{n_b}\}$ represents the skinning weights for $\mathbf{x}_{(\cdot)}$. 
Here, deformed points $\mathbf{x}_{d}$ are associated with the average of the nearest SMPL vertices' skinning weights, weighted by the point-to-point distances in deformed space. Canonical points $\mathbf{x}_{c}$ are treated analogously. 

\paragraph{Canonical Texture Representation.} The appearance is also modeled in canonical space via a neural network $f_{\text{rgb}}^{H}$ that predicts color values for 3D points $\mathbf{x}_c$ in this space. 
\begin{equation}
    f_{\text{rgb}}^{H}: \mathbb{R}^{3 + 3 + n_{\theta} + 256}  \rightarrow \mathbb{R}^{3}  .
\end{equation}

We condition the canonical texture network on the normal $\mathbf{n}_d$ in deformed space, facilitating better disentanglement of the geometry and appearance. The normals are given by the spatial gradient of the signed distance field w.r.t. the 3D location in deformed space. Following \cite{zheng2022imavatar}, the spatial gradient of the deformed shape is given by:

\begin{equation}
\begin{aligned}
    \mathbf{n}_d &= \pardev{f_{\text{sdf}}^{H}(\mathbf{x}_c, \boldsymbol{\theta})}{\mathbf{x}_d} = \pardev{f_{\text{sdf}}^{H}(\mathbf{x}_c, \boldsymbol{\theta})}{\mathbf{x}_c}  \pardev{\mathbf{x}_c}{\mathbf{x}_d} \\
    &= \pardev{f_{\text{sdf}}^{H}(\mathbf{x}_c, \boldsymbol{\theta})}{\mathbf{x}_c} (\sum_{i = 1}^{n_b} w_{d}^i \mathbf{B}_i)^{-1}.
\end{aligned}
\end{equation}
In practice we concatenate the canonical points $\mathbf{x}_c$, their normals, the pose parameters, and the extracted 256-dimensional geometry feature vectors $\textbf{z}$ from the shape network to form the input to the canonical texture network. For the remainder of this paper, we denote this neural SDF with $f_{\text{sdf}}^{H}(\mathbf{x}_{c})$ and the RGB field as $f_{\text{rgb}}^{H}(\mathbf{x}_{c})$ for brevity.

\subsection{Composited Volume Rendering}
\label{sec:volumerendering}
We extend the inverted sphere parametrization of NeRF++ \cite{kaizhang2020} to represent the scene: an outer volume (i.e., the background) covers the complement of a spherical inner volume (i.e., the space assumed to be occupied by the human) and both are modeled by separate  networks. The final pixel value is then attained via compositing.  

\paragraph{Background.}
\label{sec:bg_model}
Given the origin $\mathbf{O}$, each 3D point $\mathbf{x}_b$ = ($x_{b}, y_{b}, z_{b}$) in the outer volume is reparametrized by the quadruple $\mathbf{x}'_b$ = $(x'_{b}, y'_{b} ,z'_{b}, \frac{1}{r})$, where $\left\|\left(x'_{b}, y'_{b}, z'_{b}\right)\right\|=1$, $(x_{b}, y_{b}, z_{b}) = r \cdot (x'_{b}, y'_{b}, z'_{b})$. 
Here $r$ denotes the magnitude of the vector from the origin $\mathbf{O}$ to $\mathbf{x}_b$. This parameterization of background points leads to improved numerical stability and assigns lower resolution to farther away points. For more details, we refer to~\cite{kaizhang2020}. 
Our method is trained with videos and the background is generally not entirely static. To compensate for dynamic changes in e.g., lighting, we condition the background network $f^{B}$ on a per-frame learnable latent code $t^{i}$. 

\begin{equation}\label{eq:background}
    f^{B}: \mathbb{R}^{4 + 3 + n_{t}} \rightarrow \mathbb{R}^{1 + 3} ,
\end{equation}
where $f^{B}$ takes the 4D representation of the sampled background point $\mathbf{x}'_b$, viewing direction $\mathbf{v}$, and time encoding $t^{i}$ with dimension $n_{t}$ as input, and outputs the density and the view-dependent radiance.
\paragraph{Dynamic Foreground.}
\label{sec:fg_model}
We assume that the inner volume is occupied by a dynamic foreground -- the human we seek to reconstruct. This requires different treatment compared to~\cite{kaizhang2020} where a static foreground is modeled via a vanilla NeRF. 
In contrast, we combine the implicit neural avatar representation (\cref{sec:avatar_model}) with surface-based volume rendering~\cite{yariv2021volume}. Thus, we convert the SDF to a density $\sigma$ by applying the scaled Laplace distribution's Cumulative Distribution Function (CDF) to the negated SDF values $\xi(\mathbf{x}_c)=-f_{\text{sdf}}^{H}(\mathbf{x}_c)$:
\begin{equation}
\begin{aligned}
\sigma({\mathbf{x}_{c}})=\alpha\left(\frac{1}{2}+\frac{1}{2} \operatorname{sign}(\xi(\mathbf{x}_c)) 
(1-\exp (-\frac{|\xi(\mathbf{x}_c)|}{\beta}))\right),
\end{aligned}
\end{equation}
where $\alpha$, $\beta$ $>0$ are learnable parameters. 

Similar to \cite{yariv2021volume}, we sample $N$ points on a ray $\mathbf{r}$ = $(\mathbf{o}, \mathbf{v})$ with camera center $\mathbf{o}$ and ray direction $\mathbf{v}$ in two stages -- uniform and inverse CDF sampling. We then map the sampled points to canonical space via skeletal deformation and use standard numerical approximation to calculate the integral of the volume rendering equation:
\begin{equation}
    C^{H}(\mathbf{r}) =\sum_{i=1}^N \tau_{i} f_{\text{rgb}}^{H}(\mathbf{x}_{c}^{i})
\end{equation}
\begin{equation}
    \tau_{i} = \exp \left(-\sum_{j<i}\sigma(\mathbf{x}_{c}^{j}) \delta^{j}\right)(1 - \exp(-\sigma(\mathbf{x}_{c}^{i})\delta^{i}))
\end{equation}
where $\delta^{(i)}$ is the distance between two adjacent samples. Here, the accumulated alpha value of a pixel, which represents ray opacity, can be obtained by $\alpha^{H}(\mathbf{r}) = \sum_{i=1}^{N} \tau_{i}$.

\paragraph{Scene Composition.}
\label{sec:composition}
To attain the final pixel value for a ray $\mathbf{r}$, we raycast the human and background volumes separately, followed by a scene compositing step. Using the parameterization of the background, we sample $r$ along the ray $\mathbf{r}$ to obtain sample points in the outer volume for which we query $f^{B}$. The background component of a pixel is then given by the integrated color value $C^{B}(\mathbf{r})$ along the ray \cite{mildenhall2020nerf}. More details can be found in the \suppmat.
The final pixel color is then the composite of the foreground and background color.
\begin{equation}
    C(\mathbf{r}) = C^{H}(\mathbf{r}) + (1 - \alpha^{H}(\mathbf{r})) C^{B}(\mathbf{r}).
\end{equation}

\subsection{Scene Decomposition Objectives}
\label{sec:scenedecompobj}
Learning to decompose the scene into a dynamic human and background by simply minimizing the distance between the composited pixel value and image RGB value is still a severely ill-posed problem. This is due to the potentially moving scene, dynamic shadows, and general visual complexity. To this end, we propose two objectives that guide the optimization towards a clean and robust decoupling of the human from the background.

\paragraph{Opacity Sparseness Regularization.}
 One of the key components of our method is a loss $L_{\text{sparse}}$ to regularize the ray opacity via the dynamically updated human shape in canonical space. 
 We first warp sampled points into the canonical space and calculate the signed distance to the human shape.
 We then penalize non-zero ray opacities for rays that do not intersect with the subject. 
 This ray set is denoted as $\mathcal{R}_{\text{off}}^{i}$ for frame $i$.
\begin{equation}
\label{eq:l_cano}
\mathcal{L}_{\text{sparse}}^{i} = \frac{1}{|\mathcal{R}_{\text{off}}^{i}|} \sum_{\mathbf{r} \in \mathbf{\mathcal{R}}_{\text{off}}^{i}} |\alpha^{H}(\mathbf{r})|.
\end{equation}
Note that we conservatively update the SDF of the human shape throughout the whole training process which leads to a precise association of human and background rays.  

\paragraph{Self-supervised Ray Classification.}
Even with the shape regularization from Eq.~\ref{eq:l_cano}, we observe that the human fields still tend to model parts of the background due to the flexibility and expressive power of MLPs, especially if the subject is in contact with the scene. To further delineate dynamic foreground and background, we introduce an additional loss term to encourage ray distributions that contain either fully transparent or opaque rays:
\begin{equation}
\begin{aligned}
\mathcal{L}_{\text{BCE}}^{i} = &-\frac{1}{|\mathcal{R}^{i}|} \sum_{\mathbf{r} \in \mathbf{\mathcal{R}}^{i}} (\alpha^{H}(\mathbf{r}) \log(\alpha^{H}(\mathbf{r})) \\ &+ (1-\alpha^{H}(\mathbf{r})) \log(1-\alpha^{H}(\mathbf{r}))),
\end{aligned}
\end{equation}
where $\mathcal{R}^{i}$ denotes the sampled rays for frame $i$. This term penalizes deviations of the ray opacities from a binary $\{0,1\}$ distribution via the binary cross entropy loss. 
Intuitively this encourages the opacities to be zero for rays that hit the background and one for those that hit the human shape. 
In practice, this significantly helps separation of the subject and the background, in particular for difficult cases with similar pixel values across discontinuities. 

The final scene decomposition loss is then given by $L_{\text{dec}}$:
\begin{equation}
\mathcal{L}_{\text{dec}} = 
\lambda_{\text{BCE}} \mathcal{L}_{\text{BCE}} + \lambda_{\text{sparse}} \mathcal{L}_{\text{sparse}}.
\end{equation}

\subsection{Global Optimization}
\label{sec:globalopt}
To train the models that represent the background and the dynamic foreground jointly from videos, we formulate the training as global optimization over all frames.

\paragraph{Eikonal Loss.} Following IGR \cite{gropp2020implicit}, we leverage $\mathcal{L}_{\text{eik}}^{i} $ to force the shape network $f_{\text{sdf}}^{H}$ to satisfy the Eikonal equation in canonical space:
\begin{equation}
    \mathcal{L}_{\text{eik}}^{i} =  \mathbb{E}_{\mathbf{x}_c}\left(\|\nabla f_{\text{sdf}}^{H}(\mathbf{x}_c)\| -1\right)^{2} .
\end{equation}

\paragraph{Reconstruction Loss.} We calculate the $L_{1}$-distance between the rendered color $C(\mathbf{r})$ and the pixel's RGB value $\hat{C}(\mathbf{r})$ to attain the reconstruction loss $\mathcal{L}_{\text{rgb}}^{i}$ for frame $i$:  
\begin{equation}
\mathcal{L}_{\text{rgb}}^{i} = \frac{1}{|\mathcal{R}^{i}|} \sum_{\mathbf{r} \in \mathbf{\mathcal{R}}^{i}} |C(\mathbf{r}) - \hat{C}(\mathbf{r})| .
\end{equation}

\paragraph{Full Loss.} Given a video sequence with $F$ input frames, we minimize the combined loss function:
\begin{equation}
\begin{aligned}
\mathcal{L}(\mathbf{\Theta}) = \sum_{i=1}^{F}\mathcal{L}_{\text{rgb}}^{i}(\mathbf{\Theta}^{H}, \mathbf{\Theta}^{B})
+\lambda_{\text{dec}} \mathcal{L}_{\text{dec}}^{i}(\mathbf{\Theta}^{H})
+\lambda_{\text{eik}} \mathcal{L}_{\text{eik}}^{i}(\mathbf{\Theta}^{H}) \\
\end{aligned}
\end{equation}
where $\mathbf{\Theta}^{H}$ and $\mathbf{\Theta}^{B}$ are the sets of optimized parameters for the human and background model respectively. $\mathbf{\Theta}^{H}$ includes the shape network weights $\mathbf{\Theta}_{\text{sdf}}^{H}$, the texture network weights $\mathbf{\Theta}_{\text{rgb}}^{H}$ and per-frame pose parameters $\boldsymbol{\theta}_i$. $\mathbf{\Theta}^{B}$ contains the background density and radiance network weights.

\vspace{-0.05cm}


\newcommand{\tablemask}{
\begin{table}[t]
\small
\centering
\begin{tabular}{lccc}

\hline  Method & $\mathbf{Precision} \uparrow$ & $\mathbf{F1} \uparrow$ & $\mathbf{IoU} \uparrow$ \\
\hline
SMPL Tracking & $0.829$ & $0.781 $ & $0.659$ \\
PointRend \cite{kirillov2019pointrend}  & $0.957$ & $0.960 $ & $0.915$ \\
Ye et al. \cite{ye2022sprites} & ${0.945}$ & ${0.947}$ & $0.890$ \\
RVM \cite{rvm} & ${0.975}$ & ${0.977} $ & $0.950$ \\

\hline
w/o Scene Dec. Loss & $0.979$ & $0.974 $ & $0.942$ \\
Ours & $\mathbf{0.983}$ & $\mathbf{0.983} $ & $\mathbf{0.961}$ \\
\hline

\end{tabular}
\caption{\textbf{Quantitative evaluation on MonoPerfCap.} Our method outperforms all 2D segmentation baselines in all metrics.}
\vspace{-0.25cm}
\label{tab:mask}
\end{table}
}

\newcommand{\tablewobg}{
\begin{table}[t]
\small
\centering
\begin{tabular}{lccc}

\hline  Method & $\mathbf{IoU} \uparrow$ & $\mathbf{C}-\ell_{2} \downarrow$ & $\mathbf{N C} \uparrow$ \\
\hline

\hline
w/o BG Modeling.  & $0.811$ & $3.13 $ & $0.728$ \\
Ours & $\mathbf{0.818}$ & $\mathbf{2.66} $ & $\mathbf{0.753}$ \\
\hline

\end{tabular}
\caption{\textbf{Importance of background modeling.} Without background modeling, our method cannot recover the complete human body and the reconstruction quality is upper-bounded by the 2D segmentation module.}
\vspace{-0.25cm}
\label{tab:wobg}
\end{table}
}

\newcommand{\tablethreedpw}{
\begin{table}[t]
\small
\centering
\begin{tabular}{lccc}

\hline  Method & $\mathbf{IoU} \uparrow$ & $\mathbf{C}-\ell_{2} \downarrow$ & $\mathbf{N C} \uparrow$ \\
\hline

ICON \cite{xiu2022icon}  & $0.718$ & $3.32 $ & $0.731$ \\
SelfRecon \cite{jiang2022selfrecon}  & $0.648$ & $3.31 $ & $0.675$ \\

\hline
w/o Joint Opt.  & $0.810$ & $3.00 $ & $0.737$ \\
Ours & $\mathbf{0.818}$ & $\mathbf{2.66} $ & $\mathbf{0.753}$ \\
\hline

\end{tabular}
\caption{\textbf{Quantitative evaluation on 3DPW.} Our method consistently outperforms all baselines in all metrics (\cf \figref{fig:recon}).}
\vspace{-0.25cm}
\label{tab:threedpw}
\end{table}
}

\newcommand{\tablesynwild}{
\begin{table}[t]
\small
\centering
\begin{tabular}{lccc}

\hline  Method & $\mathbf{IoU} \uparrow$ & $\mathbf{C}-\ell_{2} \downarrow$ & $\mathbf{N C} \uparrow$ \\
\hline

ICON \cite{xiu2022icon}  & $0.764$ & $2.91 $ & $0.766$ \\
SelfRecon \cite{jiang2022selfrecon}  & $0.805$ & $2.50 $ & $0.776$ \\

\hline
  Ours & $\mathbf{0.813}$ & $\mathbf{2.35} $ & $\mathbf{0.796}$ \\
\hline

\end{tabular}
\caption{\textbf{Quantitative evaluation on SynWild.} Our method consistently outperforms all baselines in all metrics (\cf \figref{fig:recon}).}
\vspace{-0.5cm}
\label{tab:synwild}
\end{table}
}

\newcommand{\tableneuman}{
\begin{table}[t]
\small
\centering
\begin{tabular}{lccc}

\hline  Method & $\mathbf{SSIM} \uparrow$ & $\mathbf{PSNR} \uparrow $ \\ 
\hline

NeuMan \cite{jiang2022neuman}  & $0.958$ & $23.9 $ \\ 
HumanNeRF \cite{weng_humannerf_2022_cvpr}  & $0.963$ & $24.8 $ \\

\hline
  Ours & $\mathbf{0.964}$ & $\mathbf{25.1} $ \\
\hline

\end{tabular}
\caption{\textbf{Quantitative evaluation on NeuMan.} We report the quantitative results on test views. Our method achieves on-par and even better rendering quality compared to NeRF-based methods.}
\vspace{-0.5cm}
\label{tab:neuman}
\end{table}
}

\newcommand{\tablecape}{
\begin{table}[t]
\begin{tabular}{lcccc}
\hline Method & Input &$\mathbf{I o U} \uparrow$ & $\mathbf{C}-\ell_{2} \downarrow$ & $\mathbf{N C} \uparrow$ \\
\hline 
IP-Net \cite{bhatnagar2020ipnet} &3D & $0.916$ & $0.735 \mathrm{}$ & $0.843$ \\
SCANimate \cite{Saito:CVPR:2021} &3D & $\mathbf{0.941}$ & $\mathbf{0.560} \mathbf{}$ & $\mathbf{0.906}$ \\

\hline 
SCANimate \cite{Saito:CVPR:2021} &2.5D & ${0.665}$ & ${4.710} \mathrm{}$ & $0.785 $\\

Ours & 2.5D &$\mathbf{0.946}$ & $\mathbf{0.621} \mathbf{}$ & $\mathbf{0.906 }$ \\
\hline
\end{tabular}
\caption{\textbf{Quantitative evaluation on CAPE.} Our method outperforms IP-Net and SCANimate (2.5D) by a large margin and achieves comparable result with SCANimate (3D) which is trained on complete 3D meshes, a significantly easier setting compared to using partial 2.5D data as input. }

\vspace{-1em}
\label{tab:cape}
\end{table}
}

\figuredensityreg
\section{Experiments}
We first conduct ablation studies on our design choices. Next, we compare our method with state-of-the-art approaches in 2D segmentation, novel view synthesis, and reconstruction tasks. Finally, we demonstrate human reconstruction results on several in-the-wild monocular videos from different sources qualitatively.

\subsection{Datasets}
\noindent\textbf{MonoPerfCap Dataset \cite{Xu:2018:MHP:3191713.3181973}:} This dataset contains in-the-wild human performance sequences with ground-truth masks. Since our method can provide human segmentation as by-product, we use this dataset to compare our method with other off-the-shelf 2D segmentation approaches to validate the scene decomposition quality of our method.

\noindent\textbf{NeuMan Dataset \cite{jiang2022neuman}:} This dataset includes a collection of videos captured by a mobile phone, in which a single person performs walking. We use this dataset to compare our rendering quality of humans under unseen views with other approaches.

\noindent\textbf{3DPW Dataset \cite{vonMarcard2018}:} This dataset contains challenging in-the-wild video sequences along with accurate 3D human poses recovered by using IMUs and a moving camera. Moreover, it includes registered static clothed 3D human models. By animating the human model with the ground-truth poses, we can obtain quasi ground-truth scans to evaluate the surface reconstruction performance.

\noindent\textbf{SynWild Dataset:} We propose a new dataset called \textit{SynWild} for the evaluation of monocular human surface reconstruction method. We capture dynamic human subjects in a multi-view system and reconstruct the detailed geometry and texture via commercial software \cite{collet2015msft}. Then we place the textured 4D scans into realistic 3D scenes/HDRI panoramas and render monocular videos from virtual cameras, leveraging a high-quality game engine \cite{unreal}. This is the first dataset that allows for quantitative comparison in a realistic setting via semi-synthetic data.

\noindent\textbf{Evaluation Protocol:} We consider precision, F1 score, and mask IoU for human segmentation evaluation. We report volumetric IoU, Chamfer distance (cm) and normal consistency for surface reconstruction comparison. Rendering quality is measured via SSIM and PSNR.  

\figuremask
\figureneuman

\tablemask
\tableneuman

\subsection{Ablation Study}

\paragraph{Jointly Pose Optimization:} The initial pose estimate from a monocular RGB video is usually inaccurate. To evaluate the importance of jointly optimizing pose, shape, and appearance, we compare our full model to a version without jointly pose optimization. \tabref{tab:threedpw} shows that the joint optimization significantly helps in global pose alignment and to recover details (normal consistency), this is also confirmed by qualitative results. Please see the \suppmat.

\label{sec:ablation_opt}

\paragraph{Scene Decomposition Loss:}
To demonstrate the effectiveness of our proposed scene decomposition loss, we conduct an ablation experiment without this term during optimization.  Results in \tabref{tab:mask} indicate that without the scene decomposition loss, the segmentation tends to be noisy and includes parts of the background as shown in \figref{fig:densityreg}.

\subsection{2D Segmentation Comparisons} 
We generate human masks by extracting the pixels with ray opacity $\alpha^{H}(\mathbf{r})$ value of 1. Our produced masks are compared with SMPL Tracking, PointRend \cite{kirillov2019pointrend}, Ye \etal \cite{ye2022sprites} and RVM \cite{rvm} on the MonoPerfCap dataset \cite{Xu:2018:MHP:3191713.3181973}. \cite{kirillov2019pointrend} and \cite{rvm} are trained on large datasets with human-annotated masks, while \cite{ye2022sprites} rely on optical flow as motion clues to segment objects in an unsupervised manner. SMPL Tracking uses dilated projected SMPL masks as the result.
\tabref{tab:mask} shows the quantitative results. Our method consistently outperforms other baseline methods on all metrics. \figref{fig:mask} shows that other baselines struggle on the feet since there is no enough photometric contrast between the part of the shoes and the stairs. In contrast, our method is able to generate plausible human segmentation via decomposition from a 3D perspective.

\label{sec:mask}

\tablethreedpw
\tablesynwild

\subsection{View Synthesis Comparisons}
\label{sec:vs}
Though not our primary goal, we also compare with HumanNeRF \cite{weng_humannerf_2022_cvpr} and NeuMan \cite{jiang2022neuman} for the task of novel view synthesis on the NeuMan dataset. Note that both methods require additional human segmentation as input. Overall, we achieve comparable or even better performance quantitatively (\cf \tabref{tab:neuman}). As shown in \figref{fig:neuman}, NeuMan and HumanNeRF have obvious artifacts around feet and arms. This is because, a) off-the-shelf tools struggle to produce consistent masks and b) NeRF-based methods are known to have ``hazy" floaters in the space leading to visually unpleasant results. Our method produces more plausible renderings of the human with a clean separation from the background.

\subsection{Reconstruction Comparisons}
\label{sec:recon}
We compare our proposed human surface reconstruction method to several state-of-the-art approaches~\cite{xiu2022icon,jiang2022selfrecon} on both 3DPW \cite{vonMarcard2018} and SynWild. ICON (image-based)~\cite{xiu2022icon} reconstructs 3D clothed humans by learning a regression model from a large dateset of clothed human scans\cite{renderpeople}. SelfRecon (video-based)~\cite{jiang2022selfrecon} deploys implicit surface rendering to reconstruct avatars from monocular videos. Both methods rely on additional human masks as input for their methods. Despite this, our method outperforms \cite{xiu2022icon,jiang2022selfrecon} by a substantial margin on all metrics (\cf \tabref{tab:threedpw}, \tabref{tab:synwild}). The difference is more visible in qualitative comparison as shown in \figref{fig:recon}, where they tend to produce physically incorrect body reconstructions (e.g., missing arms and sunken backs).
In contrast, our method generates complete human bodies and recovers more surface details (e.g., cloth wrinkles and facial features). We attribute this to the better decoupling of humans from the background by our proposed modeling and learning schemes.

\subsection{Qualitative Results}
We demonstrate our results on several in-the-wild monocular videos from different sources: online, datasets, and self-captured video clips (\figref{fig:demo}). Our method is able to reconstruct complex cloth deformations and personalized facial features in detail. \textbf{Please refer to \suppmat for more qualitative results}.

\figurerecon
\figuredemo

\section{Conclusion}
In this paper, we present \methodname to reconstruct detailed 3D avatars from monocular in-the-wild videos via self-supervised scene decomposition. Our method does not require any groundtruth supervision or priors extracted from large datasets of clothed human scans, nor do we rely on any external segmentation modules. With carefully designed background modeling and temporally consistent canonical human representation, a global optimization with novel scene decomposition objectives is formulated to jointly optimize the parameters of the background field, the canonical human shape and appearance, and the human pose estimates over the entire sequence via differentiable composited volume rendering. Our method achieves robust and high-fidelity human reconstruction from monocular videos. 

\noindent\textbf{Limitations:} Although readily available, \methodname relies on reasonable pose estimates as inputs. Furthermore, loose clothing such as skirts or free-flowing garments poses significant challenges due to their fast dynamics. We refer to \suppmat for a more detailed discussion of limitations and societal impact.

{\small
\bibliographystyle{ieee_fullname}
\bibliography{main}

\begin{thebibliography}{10}\itemsep=-1pt

\bibitem{renderpeople}
{\em Renderpeople}, 2018.
\newblock \url{https://www.renderpeople.com}.

\bibitem{unreal}
{\em Unreal}, 2020.
\newblock \url{https://www.unrealengine.com}.

\bibitem{alldieck2018video}
Thiemo Alldieck, Marcus Magnor, Weipeng Xu, Christian Theobalt, and Gerard
  Pons-Moll.
\newblock Video based reconstruction of 3d people models.
\newblock In {\em Proceedings of the IEEE Conference on Computer Vision and
  Pattern Recognition}, pages 8387--8397, 2018.

\bibitem{alldieck2022phorhum}
Thiemo Alldieck, Mihai Zanfir, and Cristian Sminchisescu.
\newblock Photorealistic monocular 3d reconstruction of humans wearing
  clothing.
\newblock In {\em Proceedings of the IEEE/CVF Conference on Computer Vision and
  Pattern Recognition (CVPR)}, 2022.

\bibitem{10.1007/978-3-319-46484-8_26}
Pia Bideau and Erik Learned-Miller.
\newblock It's moving! a probabilistic model for causal motion segmentation in
  moving camera videos.
\newblock In Bastian Leibe, Jiri Matas, Nicu Sebe, and Max Welling, editors,
  {\em Computer Vision -- ECCV 2016}, pages 433--449, Cham, 2016. Springer
  International Publishing.

\bibitem{Bozic_2021_CVPR}
Aljaz Bozic, Pablo Palafox, Michael Zollhofer, Justus Thies, Angela Dai, and
  Matthias Niessner.
\newblock Neural deformation graphs for globally-consistent non-rigid
  reconstruction.
\newblock In {\em Proceedings of the IEEE/CVF Conference on Computer Vision and
  Pattern Recognition (CVPR)}, pages 1450--1459, June 2021.

\bibitem{burov2021dsfn}
Andrei Burov, Matthias Nie{\ss}ner, and Justus Thies.
\newblock Dynamic surface function networks for clothed human bodies.
\newblock In {\em Proc. International Conference on Computer Vision (ICCV)},
  Oct. 2021.

\bibitem{casado2022pergamo}
Andrés Casado-Elvira, Marc Comino~Trinidad, and Dan Casas.
\newblock {PERGAMO}: Personalized 3d garments from monocular video.
\newblock {\em Computer Graphics Forum (Proc. of SCA), 2022}, 2022.

\bibitem{collet2015msft}
Alvaro Collet, Ming Chuang, Pat Sweeney, Don Gillett, Dennis Evseev, David
  Calabrese, Hugues Hoppe, Adam Kirk, and Steve Sullivan.
\newblock High-quality streamable free-viewpoint video.
\newblock {\em ACM Trans. Graph.}, 34(4), jul 2015.

\bibitem{10.1145/1360612.1360697}
Edilson de Aguiar, Carsten Stoll, Christian Theobalt, Naveed Ahmed, Hans-Peter
  Seidel, and Sebastian Thrun.
\newblock Performance capture from sparse multi-view video.
\newblock 27(3):1–10, 2008.

\bibitem{Dong_2022_CVPR}
Zijian Dong, Chen Guo, Jie Song, Xu Chen, Andreas Geiger, and Otmar Hilliges.
\newblock Pina: Learning a personalized implicit neural avatar from a single
  rgb-d video sequence.
\newblock In {\em Proceedings of the IEEE/CVF Conference on Computer Vision and
  Pattern Recognition (CVPR)}, pages 20470--20480, June 2022.

\bibitem{li2022neurips}
Qiao Feng, Yebin Liu, Yu-Kun Lai, Jingyu Yang, and Kun Li.
\newblock Fof: Learning fourier occupancy field for monocular real-time human
  reconstruction.
\newblock In {\em NeurIPS}, 2022.

\bibitem{gropp2020implicit}
Amos Gropp, Lior Yariv, Niv Haim, Matan Atzmon, and Yaron Lipman.
\newblock Implicit geometric regularization for learning shapes.
\newblock {\em arXiv preprint arXiv:2002.10099}, 2020.

\bibitem{guo2021human}
Chen Guo, Xu Chen, Jie Song, and Otmar Hilliges.
\newblock Human performance capture from monocular video in the wild.
\newblock In {\em 2021 International Conference on 3D Vision (3DV)}, pages
  889--898. IEEE, 2021.

\bibitem{deepcap}
Marc Habermann, Weipeng Xu, Michael Zollhoefer, Gerard Pons-Moll, and Christian
  Theobalt.
\newblock Deepcap: Monocular human performance capture using weak supervision.
\newblock In {\em {IEEE} Conference on Computer Vision and Pattern Recognition
  (CVPR)}. {IEEE}, jun 2020.

\bibitem{10.1145/3311970}
Marc Habermann, Weipeng Xu, Michael Zollh\"{o}fer, Gerard Pons-Moll, and
  Christian Theobalt.
\newblock Livecap: Real-time human performance capture from monocular video.
\newblock {\em ACM Trans. Graph.}, 38(2), mar 2019.

\bibitem{NEURIPS2020_690f44c8}
Tong He, John Collomosse, Hailin Jin, and Stefano Soatto.
\newblock Geo-pifu: Geometry and pixel aligned implicit functions for
  single-view human reconstruction.
\newblock In H. Larochelle, M. Ranzato, R. Hadsell, M.F. Balcan, and H. Lin,
  editors, {\em Advances in Neural Information Processing Systems}, volume~33,
  pages 9276--9287. Curran Associates, Inc., 2020.

\bibitem{He_2021_ICCV}
Tong He, Yuanlu Xu, Shunsuke Saito, Stefano Soatto, and Tony Tung.
\newblock Arch++: Animation-ready clothed human reconstruction revisited.
\newblock In {\em Proceedings of the IEEE/CVF International Conference on
  Computer Vision (ICCV)}, pages 11046--11056, October 2021.

\bibitem{1335229}
A. Hilton and J. Starck.
\newblock Multiple view reconstruction of people.
\newblock In {\em Proceedings. 2nd International Symposium on 3D Data
  Processing, Visualization and Transmission, 2004. 3DPVT 2004.}, pages
  357--364, 2004.

\bibitem{HVTR:3DV2022}
Tao Hu, Tao Yu, Zerong Zheng, He Zhang, Yebin Liu, and Matthias Zwicker.
\newblock Hvtr: Hybrid volumetric-textural rendering for human avatars.
\newblock In {\em 2022 International Conference on 3D Vision (3DV)}, 2022.

\bibitem{huang2020arch}
Zeng Huang, Yuanlu Xu, Christoph Lassner, Hao Li, and Tony Tung.
\newblock Arch: Animatable reconstruction of clothed humans.
\newblock In {\em Proceedings of the IEEE/CVF Conference on Computer Vision and
  Pattern Recognition}, pages 3093--3102, 2020.

\bibitem{zhang2021stnerf}
Zhang Jiakai, Liu Xinhang, Ye Xinyi, Zhao Fuqiang, Zhang Yanshun, Wu Minye,
  Zhang Yingliang, Xu Lan, and Yu Jingyi.
\newblock Editable free-viewpoint video using a layered neural representation.
\newblock In {\em ACM SIGGRAPH}, 2021.

\bibitem{jiang2022selfrecon}
Boyi Jiang, Yang Hong, Hujun Bao, and Juyong Zhang.
\newblock Selfrecon: Self reconstruction your digital avatar from monocular
  video.
\newblock In {\em {IEEE/CVF} Conference on Computer Vision and Pattern
  Recognition (CVPR)}, 2022.

\bibitem{jiang2022neuman}
Wei Jiang, Kwang~Moo Yi, Golnoosh Samei, Oncel Tuzel, and Anurag Ranjan.
\newblock Neuman: Neural human radiance field from a single video.
\newblock In {\em Proceedings of the European conference on computer vision
  (ECCV)}, 2022.

\bibitem{MODNet}
Zhanghan Ke, Jiayu Sun, Kaican Li, Qiong Yan, and Rynson~W.H. Lau.
\newblock Modnet: Real-time trimap-free portrait matting via objective
  decomposition.
\newblock In {\em AAAI}, 2022.

\bibitem{kirillov2019pointrend}
Alexander Kirillov, Yuxin Wu, Kaiming He, and Ross Girshick.
\newblock {PointRend}: Image segmentation as rendering.
\newblock 2019.

\bibitem{leroy:hal-02975479}
Vincent Leroy, Jean-S{\'e}bastien Franco, and Edmond Boyer.
\newblock {Volume Sweeping: Learning Photoconsistency for Multi-View Shape
  Reconstruction}.
\newblock {\em {International Journal of Computer Vision}}, 129:284--299, Feb.
  2021.

\bibitem{li2022tava}
Ruilong Li, Julian Tanke, Minh Vo, Michael Zollhofer, Jurgen Gall, Angjoo
  Kanazawa, and Christoph Lassner.
\newblock Tava: Template-free animatable volumetric actors.
\newblock 2022.

\bibitem{li2021posefusion}
Zhe Li, Tao Yu, Zerong Zheng, Kaiwen Guo, and Yebin Liu.
\newblock Posefusion: Pose-guided selective fusion for single-view human
  volumetric capture.
\newblock In {\em Proceedings of the IEEE/CVF Conference on Computer Vision and
  Pattern Recognition}, pages 14162--14172, 2021.

\bibitem{rvm}
Shanchuan Lin, Linjie Yang, Imran Saleemi, and Soumyadip Sengupta.
\newblock Robust high-resolution video matting with temporal guidance, 2021.

\bibitem{journals/tvcg/LiuDX10}
Yebin Liu, Qionghai Dai, and Wenli Xu.
\newblock A point-cloud-based multiview stereo algorithm for free-viewpoint
  video.
\newblock {\em IEEE Trans. Vis. Comput. Graph.}, 16(3):407--418, 2010.

\bibitem{loper2015smpl}
Matthew Loper, Naureen Mahmood, Javier Romero, Gerard Pons-Moll, and Michael~J
  Black.
\newblock Smpl: A skinned multi-person linear model.
\newblock {\em ACM transactions on graphics (TOG)}, 34(6):1--16, 2015.

\bibitem{mildenhall2020nerf}
Ben Mildenhall, Pratul~P Srinivasan, Matthew Tancik, Jonathan~T Barron, Ravi
  Ramamoorthi, and Ren Ng.
\newblock Nerf: Representing scenes as neural radiance fields for view
  synthesis.
\newblock In {\em European conference on computer vision}, pages 405--421.
  Springer, 2020.

\bibitem{Moon_2022_ECCV_ClothWild}
Gyeongsik Moon, Hyeongjin Nam, Takaaki Shiratori, and Kyoung~Mu Lee.
\newblock 3d clothed human reconstruction in the wild.
\newblock In {\em European Conference on Computer Vision (ECCV)}, 2022.

\bibitem{newcombe2015dynamicfusion}
Richard~A Newcombe, Dieter Fox, and Steven~M Seitz.
\newblock Dynamicfusion: Reconstruction and tracking of non-rigid scenes in
  real-time.
\newblock In {\em Proceedings of the IEEE conference on computer vision and
  pattern recognition}, pages 343--352, 2015.

\bibitem{newcombe2011kinectfusion}
Richard~A Newcombe, Shahram Izadi, Otmar Hilliges, David Molyneaux, David Kim,
  Andrew~J Davison, Pushmeet Kohi, Jamie Shotton, Steve Hodges, and Andrew
  Fitzgibbon.
\newblock Kinectfusion: Real-time dense surface mapping and tracking.
\newblock In {\em 2011 10th IEEE international symposium on mixed and augmented
  reality}, pages 127--136. IEEE, 2011.

\bibitem{2021narf}
Atsuhiro Noguchi, Xiao Sun, Stephen Lin, and Tatsuya Harada.
\newblock Neural articulated radiance field.
\newblock In {\em International Conference on Computer Vision}, 2021.

\bibitem{6751331}
Anestis Papazoglou and Vittorio Ferrari.
\newblock Fast object segmentation in unconstrained video.
\newblock In {\em 2013 IEEE International Conference on Computer Vision}, pages
  1777--1784, 2013.

\bibitem{peng2021animatable}
Sida Peng, Junting Dong, Qianqian Wang, Shangzhan Zhang, Qing Shuai, Xiaowei
  Zhou, and Hujun Bao.
\newblock Animatable neural radiance fields for modeling dynamic human bodies.
\newblock In {\em ICCV}, 2021.

\bibitem{peng2021neural}
Sida Peng, Yuanqing Zhang, Yinghao Xu, Qianqian Wang, Qing Shuai, Hujun Bao,
  and Xiaowei Zhou.
\newblock Neural body: Implicit neural representations with structured latent
  codes for novel view synthesis of dynamic humans.
\newblock In {\em Proceedings of the IEEE/CVF Conference on Computer Vision and
  Pattern Recognition}, pages 9054--9063, 2021.

\bibitem{saito2019pifu}
Shunsuke Saito, Zeng Huang, Ryota Natsume, Shigeo Morishima, Angjoo Kanazawa,
  and Hao Li.
\newblock Pifu: Pixel-aligned implicit function for high-resolution clothed
  human digitization.
\newblock In {\em Proceedings of the IEEE/CVF International Conference on
  Computer Vision}, pages 2304--2314, 2019.

\bibitem{saito2020pifuhd}
Shunsuke Saito, Tomas Simon, Jason Saragih, and Hanbyul Joo.
\newblock Pifuhd: Multi-level pixel-aligned implicit function for
  high-resolution 3d human digitization.
\newblock In {\em Proceedings of the IEEE/CVF Conference on Computer Vision and
  Pattern Recognition}, pages 84--93, 2020.

\bibitem{BMSengupta20}
Soumyadip Sengupta, Vivek Jayaram, Brian Curless, Steve Seitz, and Ira
  Kemelmacher-Shlizerman.
\newblock Background matting: The world is your green screen.
\newblock In {\em Computer Vision and Pattern Regognition (CVPR)}, 2020.

\bibitem{https://doi.org/10.48550/arxiv.2207.11232}
Prafull Sharma, Ayush Tewari, Yilun Du, Sergey Zakharov, Rares Ambrus, Adrien
  Gaidon, William~T. Freeman, Fredo Durand, Joshua~B. Tenenbaum, and Vincent
  Sitzmann.
\newblock Seeing 3d objects in a single image via self-supervised
  static-dynamic disentanglement, 2022.

\bibitem{4178157}
Jonathan Starck and Adrian Hilton.
\newblock Surface capture for performance-based animation.
\newblock {\em IEEE Computer Graphics and Applications}, 27(3):21--31, 2007.

\bibitem{su2022danbo}
Shih-Yang Su, Timur Bagautdinov, and Helge Rhodin.
\newblock Danbo: Disentangled articulated neural body representations via graph
  neural networks.
\newblock In {\em European Conference on Computer Vision}, 2022.

\bibitem{su2021anerf}
Shih-Yang Su, Frank Yu, Michael Zollh{\"o}fer, and Helge Rhodin.
\newblock A-nerf: Articulated neural radiance fields for learning human shape,
  appearance, and pose.
\newblock In {\em Advances in Neural Information Processing Systems}, 2021.

\bibitem{tschernezki21neuraldiff}
Vadim Tschernezki, Diane Larlus, and Andrea Vedaldi.
\newblock {NeuralDiff}: Segmenting {3D} objects that move in egocentric videos.
\newblock In {\em Proceedings of the International Conference on {3D} Vision
  (3DV)}, 2021.

\bibitem{tsiminaki:hal-00977755}
Vagia Tsiminaki, Jean-S{\'e}bastien Franco, and Edmond Boyer.
\newblock {High Resolution 3D Shape Texture from Multiple Videos}.
\newblock In {\em {CVPR 2014 - IEEE International Conference on Computer Vision
  and Pattern Recognition}}, pages 1502--1509, Columbus, OH, United States,
  June 2014. {IEEE}.

\bibitem{vonMarcard2018}
Timo von Marcard, Roberto Henschel, Michael Black, Bodo Rosenhahn, and Gerard
  Pons-Moll.
\newblock Recovering accurate 3d human pose in the wild using imus and a moving
  camera.
\newblock In {\em European Conference on Computer Vision (ECCV)}, sep 2018.

\bibitem{ARAH:2022:ECCV}
Shaofei Wang, Katja Schwarz, Andreas Geiger, and Siyu Tang.
\newblock Arah: Animatable volume rendering of articulated human sdfs.
\newblock In {\em European Conference on Computer Vision}, 2022.

\bibitem{weng_humannerf_2022_cvpr}
Chung-Yi Weng, Brian Curless, Pratul~P. Srinivasan, Jonathan~T. Barron, and Ira
  Kemelmacher-Shlizerman.
\newblock Human{N}e{RF}: Free-viewpoint rendering of moving people from
  monocular video.
\newblock In {\em Proceedings of the IEEE/CVF Conference on Computer Vision and
  Pattern Recognition (CVPR)}, pages 16210--16220, June 2022.

\bibitem{https://doi.org/10.48550/arxiv.2205.15838}
Tianhao Wu, Fangcheng Zhong, Andrea Tagliasacchi, Forrester Cole, and Cengiz
  Oztireli.
\newblock D$^2$nerf: Self-supervised decoupling of dynamic and static objects
  from a monocular video, 2022.

\bibitem{8953201}
Christopher Xie, Yu Xiang, Zaid Harchaoui, and Dieter Fox.
\newblock Object discovery in videos as foreground motion clustering.
\newblock In {\em 2019 IEEE/CVF Conference on Computer Vision and Pattern
  Recognition (CVPR)}, pages 9986--9995, 2019.

\bibitem{xiu2022icon}
Yuliang Xiu, Jinlong Yang, Dimitrios Tzionas, and Michael~J. Black.
\newblock {ICON}: {I}mplicit {C}lothed humans {O}btained from {N}ormals.
\newblock In {\em Proceedings of the IEEE/CVF Conference on Computer Vision and
  Pattern Recognition (CVPR)}, pages 13296--13306, June 2022.

\bibitem{xu2021hnerf}
Hongyi Xu, Thiemo Alldieck, and Cristian Sminchisescu.
\newblock H-ne{RF}: Neural radiance fields for rendering and temporal
  reconstruction of humans in motion.
\newblock In A. Beygelzimer, Y. Dauphin, P. Liang, and J.~Wortman Vaughan,
  editors, {\em Advances in Neural Information Processing Systems}, 2021.

\bibitem{xu2022sanerf}
Tianhan Xu, Yasuhiro Fujita, and Eiichi Matsumoto.
\newblock Surface-aligned neural radiance fields for controllable 3d human
  synthesis.
\newblock In {\em CVPR}, 2022.

\bibitem{Xu:2018:MHP:3191713.3181973}
Weipeng Xu, Avishek Chatterjee, Michael Zollh\"{o}fer, Helge Rhodin, Dushyant
  Mehta, Hans-Peter Seidel, and Christian Theobalt.
\newblock Monoperfcap: Human performance capture from monocular video.
\newblock 37(2):27:1--27:15, May 2018.

\bibitem{yang2021selfsupervised}
Charig Yang, Hala Lamdouar, Erika Lu, Andrew Zisserman, and Weidi Xie.
\newblock Self-supervised video object segmentation by motion grouping.
\newblock In {\em ICCV}, 2021.

\bibitem{yariv2021volume}
Lior Yariv, Jiatao Gu, Yoni Kasten, and Yaron Lipman.
\newblock Volume rendering of neural implicit surfaces.
\newblock In {\em Thirty-Fifth Conference on Neural Information Processing
  Systems}, 2021.

\bibitem{yariv2020multiview}
Lior Yariv, Yoni Kasten, Dror Moran, Meirav Galun, Matan Atzmon, Basri Ronen,
  and Yaron Lipman.
\newblock Multiview neural surface reconstruction by disentangling geometry and
  appearance.
\newblock {\em Advances in Neural Information Processing Systems}, 33, 2020.

\bibitem{ye2022sprites}
Vickie Ye, Zhengqi Li, Richard Tucker, Angjoo Kanazawa, and Noah Snavely.
\newblock Deformable sprites for unsupervised video decomposition.
\newblock In {\em IEEE Conference on Computer Vision and Pattern Recognition
  (CVPR)}, June 2022.

\bibitem{BodyFusion}
Tao Yu, Kaiwen Guo, Feng Xu, Yuan Dong, Zhaoqi Su, Jianhui Zhao, Jianguo Li,
  Qionghai Dai, and Yebin Liu.
\newblock Bodyfusion: Real-time capture of human motion and surface geometry
  using a single depth camera.
\newblock In {\em The IEEE International Conference on Computer Vision (ICCV)}.
  IEEE, October 2017.

\bibitem{DoubleFusion}
Tao Yu, Zerong Zheng, Kaiwen Guo, Jianhui Zhao, Qionghai Dai, Hao Li, Gerard
  Pons-Moll, and Yebin Liu.
\newblock Doublefusion: Real-time capture of human performances with inner body
  shapes from a single depth sensor.
\newblock In {\em The IEEE International Conference on Computer Vision and
  Pattern Recognition(CVPR)}. IEEE, June 2018.

\bibitem{yuan2021star}
Wentao Yuan, Zhaoyang Lv, Tanner Schmidt, and Steven Lovegrove.
\newblock Star: Self-supervised tracking and reconstruction of rigid objects in
  motion with neural rendering.
\newblock In {\em Proceedings of the IEEE/CVF Conference on Computer Vision and
  Pattern Recognition}, pages 13144--13152, 2021.

\bibitem{kaizhang2020}
Kai Zhang, Gernot Riegler, Noah Snavely, and Vladlen Koltun.
\newblock Nerf++: Analyzing and improving neural radiance fields.
\newblock {\em arXiv:2010.07492}, 2020.

\bibitem{zheng2022imavatar}
Yufeng Zheng, Victoria~Fernández Abrevaya, Marcel~C. Bühler, Xu Chen,
  Michael~J. Black, and Otmar Hilliges.
\newblock {I} {M} {Avatar}: Implicit morphable head avatars from videos.
\newblock In {\em Computer Vision and Pattern Recognition (CVPR)}, 2022.

\bibitem{zheng2021pamir}
Zerong Zheng, Tao Yu, Yebin Liu, and Qionghai Dai.
\newblock Pamir: Parametric model-conditioned implicit representation for
  image-based human reconstruction.
\newblock {\em IEEE Transactions on Pattern Analysis and Machine Intelligence},
  2021.

\end{thebibliography}
}

\end{document}